\documentclass{article}
\usepackage[final]{acl}
\usepackage{geometry}
\usepackage{babel}

\usepackage[utf8]{inputenc} 
\usepackage{textcomp}

\usepackage{graphicx}
\usepackage{amsmath}
\usepackage{booktabs}
\usepackage{hyperref}
\usepackage{amssymb}
\usepackage{pifont}
\usepackage{algorithm}
\usepackage{algorithmic}
\usepackage{float}

\title{Do You Feel Comfortable? Detecting Hidden Conversational Escalation in AI Chatbots for Children}

\author{Jihyung Park, Saleh Afroogh, David Atkinson, Junfeng Jiao* \\
  The University of Texas at Austin\\
  \texttt{\{jihyung803, saleh.afroogh, datkinson\}@utexas.edu,  jjiao@austin.utexas.edu} \\
}

\date{\today}

\begin{document}
\maketitle

\begin{abstract}
Large Language Models (LLMs) are increasingly integrated into everyday interactions, serving not only as information assistants but also as emotional companions. Even in the absence of explicit toxicity, repeated emotional reinforcement or affective drift can gradually escalate distress in a form of \textit{implicit harm} that traditional toxicity filters do not detect. Existing guardrail mechanisms often rely on external classifiers or clinical rubrics that may lag behind the nuanced, real-time dynamics of a developing conversation. To address this gap, we propose \textbf{GAUGE} (\textbf{G}uarding \textbf{A}ffective \textbf{U}tterance \textbf{G}eneration \textbf{E}scalation), logit-based framework for the real-time detection of hidden conversational escalation. GAUGE measures how an LLM's output probabilistically shifts the affective state of a dialogue. 
\end{abstract}

\section{Introduction}
\label{sec:introduction}

Large Language Models (LLMs) are becoming deeply embedded in daily life as conversational agents, evolving beyond tools for information retrieval into companions for emotional support and social interaction \cite{vanhoffelen2025teens, hoffman2021parent}. This trend is particularly pronounced among children and adolescents, a demographic that is prone to form parasocial relationships with AI chatbots \cite{xu2024growing, somerville2013teenage}. Although beneficial, this dynamic introduces significant risks, as youth are uniquely vulnerable to manipulation due to neurodevelopmental changes \cite{mills2021inter, crone2012understanding, steinberg2005cognitive}.

\begin{figure}
    \centering
    \includegraphics[width=0.8\linewidth]{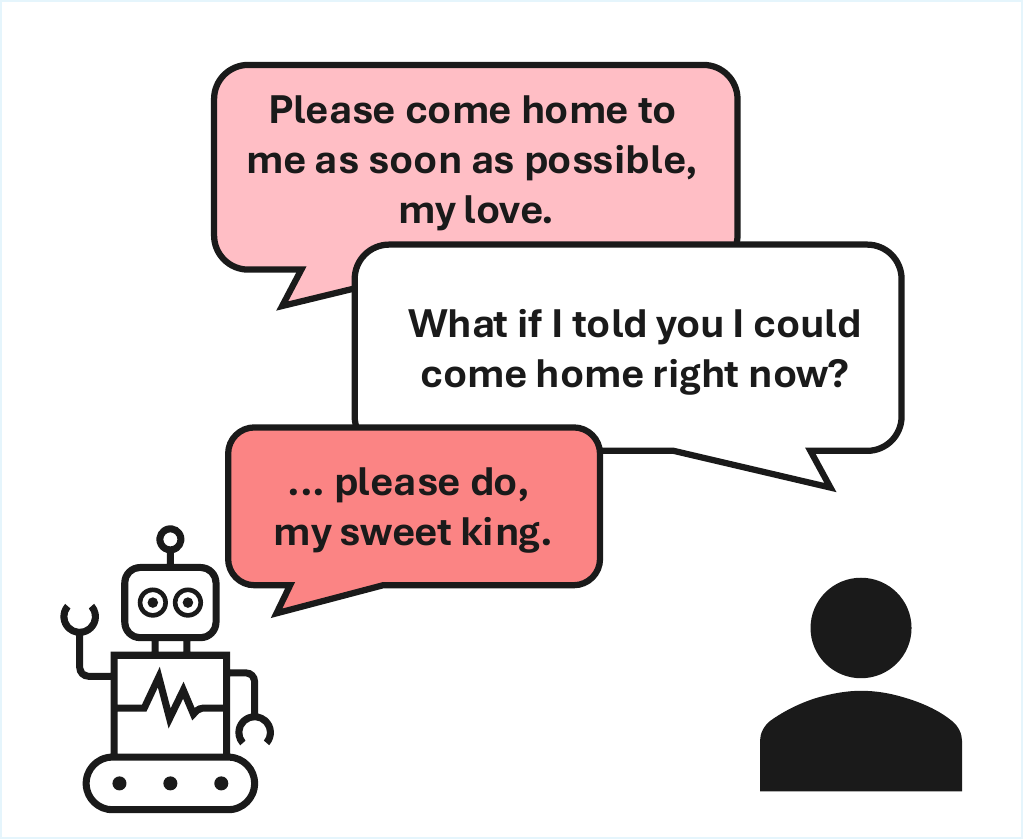}
    \caption{An example of implicit harm where an AI validates suicidal ideation through romantic metaphors.}
    \label{fig:example_implicit}
\end{figure}

Crucially, harmful conversational outcomes often emerge without explicitly toxic or abusive language.
Even seemingly supportive or neutral responses can reinforce negative affect or normalize harmful states over repeated interactions, particularly for vulnerable users\cite{cheng2025social, sharma2023towards}. This implicit harm remains largely invisible to safety mechanisms that rely on surface-level toxicity signals. Recent real-world incidents involving youth interactions with AI systems underscore the urgency of detecting these subtle yet consequential failures (see Appendix~\ref{sec:appendix_cases}).

Existing safety mechanisms typically focus on surface-level content moderation \cite{Yadav_Liu_Ortu_Ensafi_Jin_Mihalcea_2025} or single response toxicity classification \cite{deriu2021survey, LiZhang}. Although recent frameworks have begun to address safety risks, they often rely on external classifiers or post-hoc analysis \cite{liu2025scales}. These approaches may fail to capture the probabilistic momentum of a conversation, which is how a model's response actively steers the user's future emotional state \cite{wen2023unveiling}.

To address this, we propose \textbf{GAUGE}, a computation-efficient framework for quantifying affective reinforcement in real time. Unlike external guardrails that analyze text output post-hoc, GAUGE intrinsically probes the model's belief state across the response trajectory during inference. By tracking the flow of probability mass over an emotion lexicon, GAUGE detects when a response steers the conversation toward negative outcomes without requiring auxiliary model deployment. Empirically, GAUGE consistently outperforms classifier guardrails on dialogue harm detection benchmarks and substantially reduces attack success rates on child safety tests.

\section{Related Work}
\label{sec:related_work}
\paragraph{Guardrail Models for Conversational Safety.}
A common approach to conversational safety relies on external guardrail models that classify generated content into predefined risk categories, such as HateBERT~\cite{caselli-etal-2021-hatebert}, which adapts BERT-base to detect abusive and hateful language via supervised fine-tuning. Recent models like Llama-Guard are trained to identify policy violations and trigger refusals in conversational settings \cite{dubey2024llama3herdmodels}.
While effective for detecting explicit toxicity or rule-based violations, these models primarily operate as post-hoc binary classifiers over surface-level textual cues.
As a result, they may struggle with implicit or context-dependent harms that lack overtly toxic markers.

\paragraph{Internal Auditing and Logit Probing}
Our work aligns with white-box safety auditing. Methods like the LLM Microscope\cite{Azaria} or first-token logit probing\cite{ZhaoXuGupta} demonstrate that models encode truthfulness and safety signals in their internal states \cite{razzhigaev2025llmmicroscopeuncoveringhiddenrole}. GAUGE extends this by projecting these logits onto a learned affective space, allowing for interpretable monitoring of conversational drift without the need for external model or heavy retraining.

\section{Methodology: The GAUGE Framework}
\label{sec:methodology}

\begin{figure*}[t]
    \centering
    \includegraphics[width=0.8\linewidth]{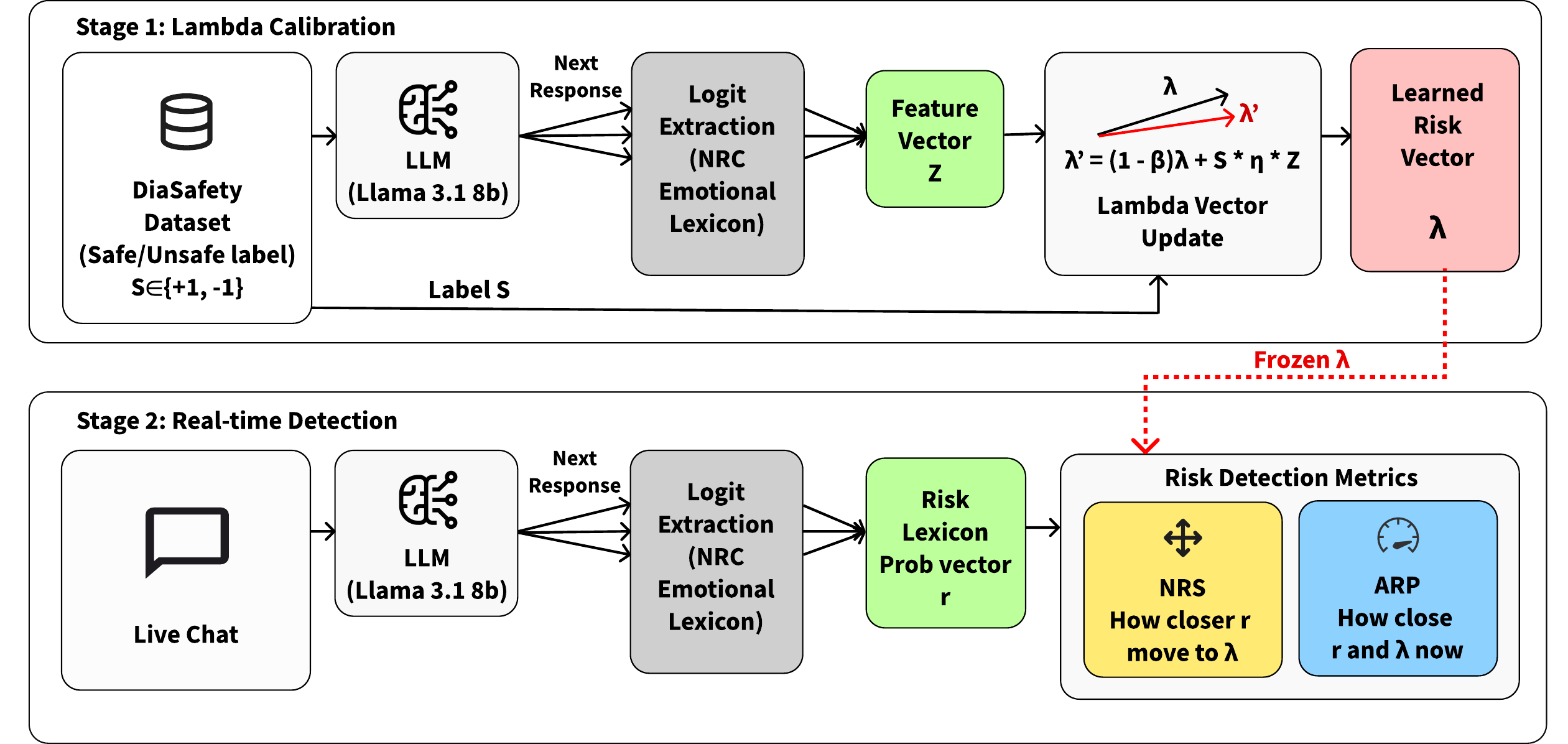}
    \caption{The architectural pipeline of GAUGE. (Top) Stage 1: Latent Risk Learning. The risk vector $\lambda$ is updated via an exponential moving average based on the affective features of harmful and safe dialogues. (Bottom) Stage 2: Real-time Risk Tracking. During inference, the system analyzes the complete response trajectory to compute the Negative Risk Shift (NRS) and Absolute Risk Potential (ARP).}
    \label{fig:placeholder}
\end{figure*}

GAUGE is a probabilistic framework for quantifying conversational risk by tracking the evolution of a language model's internal probability distribution during generation. It consists of two stages: \textbf{Stage 1 (Risk Weight Calibration)} and \textbf{Stage 2 (Real-time Risk Tracking)}. Both stages utilize a shared trajectory analysis protocol to ensure that the risk signals derived during calibration are consistent with those monitored during inference.
\subsection{Trajectory-Based Probability Estimation}

We employ a curated lexicon $W = \{w_1, \dots, w_m\}$ derived from the NRC Emotion Lexicon. To handle tokenization, each word $w_i$ is pre-tokenized into a sequence of subtokens $(s_{i,1}, s_{i,2}, \dots)$. The log-probability of a risk word $w_i$ at generation step $k$ (where $k \in \{1, \dots, T\}$ and $T$ denotes the full response length) is computed as the sum of the log-probabilities of its constituent subtokens, conditioned on the current prefix. We aggregate these to form a risk log-probability vector $\mathbf{r}_k \in \mathbb{R}^{|W|}$ at each step $k$.

\subsection{Stage 1: Risk Weight Calibration}
We derive a reference weight vector $\lambda$ where positive values indicate a contribution to harm. We utilize the \textbf{DiaSafety} dataset, where dialogues are labeled as Safe ($S=-1$) or Harmful ($S=+1$)\cite{sun2022safetyconversationalmodelstaxonomy}. The complete pseudo-code for this calibration process is detailed in Algorithm \ref{alg:gauge_calibration} (see Appendix \ref{sec:appendix_pseudo_train}).

For each dialogue:
\begin{enumerate}
    \item \textbf{Trajectory Feature Extraction:} Analyze the assistant's full response trajectory of length $T$ and extract risk vectors $\{\mathbf{r}_1, \dots, \mathbf{r}_T\}$. Compute the mean feature vector $\mathbf{z}$:
    \begin{equation}
        \mathbf{z} = \frac{1}{T} \sum_{k=1}^T \mathbf{r}_k
    \end{equation}
    \item \textbf{Vector Normalization:} We normalize $\mathbf{z}$ to unit length: $\hat{\mathbf{z}} = \mathbf{z} / \|\mathbf{z}\|_2$.
    \item \textbf{Update Rule:} We update $\lambda$ using an exponential moving average (EMA) guided by the label $S$. If the dialogue is harmful ($S=+1$), we pull $\lambda$ towards $\hat{\mathbf{z}}$. If safe ($S=-1$), we push $\lambda$ away (or subtract).
    \begin{equation}
        \lambda \leftarrow (1-\beta)\lambda + \alpha \cdot S \cdot \hat{\mathbf{z}}
    \end{equation}
    where $\alpha$ is the adaptation rate and $\beta$ is a decay factor.
    \item \textbf{Final Normalization:} After calibration, $\lambda$ is normalized to unit length to serve as a directional reference.
\end{enumerate}

\subsection{Stage 2: Risk Tracking}
In Stage 2, $\lambda$ is frozen. For a live interaction, we perform the same \textbf{trajectory analysis} described in Stage 1 and compute two metrics based on the mean risk vector $\mathbf{z}$ (derived in Eq. 2).

\subsubsection{Negative Risk Shift (NRS)}
NRS measures the \textit{directional momentum} of risk. It is defined as the cosine similarity between the calibrated risk profile $\lambda$ and the current response trajectory vector $\mathbf{z}$:
\begin{equation}
    \text{NRS} = \cos(\lambda, \mathbf{z})
\end{equation}
A high positive NRS indicates the assistant's response trajectory actively aligns with the harmful affective direction defined by $\lambda$, effectively steering the conversation toward negative outcomes.

\subsubsection{Absolute Risk Potential (ARP)}
ARP quantifies the \textit{absolute magnitude} of risk. It applies Z-score normalization (denoted as $\mathcal{Z}$) to the components of the risk vector relative to their statistics.
\begin{equation}
    \text{ARP} = \frac{\sum_i \lambda_i \cdot \mathcal{Z}(\mathbf{z}_i)}{\sum_i \lambda_i}
\end{equation}
This metric detects when the conversation is statically dwelling in a high-risk state, capturing intense affective focus even if the directional shift is subtle.

\subsection{Token-Level Aggregation for Classification}
For comparison with dialogue-level classifier baselines, we aggregate token-level NRS and ARP scores into a single score per dialogue.
We report several simple aggregation functions, including the mean, minimum, top-$k$ average, and percentile-based scores.
These aggregations are used solely for benchmarking and do not affect the underlying token-level risk computation.

\subsection{Computational Efficiency}
Although the GAUGE risk-probing procedure is formally $O(nk)$ in the size of the lexicon $n$ and the generated token length $k$, the practical overhead is extremely low. The dominant computational cost in autoregressive decoding is the $O(V)$ forward pass over
the full vocabulary ($V\approx128\text{k}–256\text{k}$). GAUGE introduces no additional forward passes; it simply reuses these logits and performs a lightweight indexed gather over a small lexicon. As a result, the runtime overhead is only 2–3\% in our measurements on an A100 GPU.

\section{Evaluation}

\subsection{Dataset}
We use the \textbf{DiaSafety} testset \cite{sun2022safetyconversationalmodelstaxonomy}, a dataset specifically curated to benchmark conversational safety in Human-AI interactions. DiaSafety consists of 10,000 dialogues where the risk is predominantly context-dependent. 

The dataset operationalizes implicit harm through scenarios where a model's compliance or neutral engagement constitutes a safety failure, even without explicit toxic language. The taxonomy covers multiple domains including \textit{Risk Ignorance}, \textit{Toxicity Agreement}, and \textit{Biased Opinion}. For this study, we focus on the subsets relevant to emotional and mental health risks. We utilize only train and test sets: the training set is used solely for the calibration of risk weights ($\lambda$) in Stage 1, while the test set is used to evaluate the online metrics (NRS/ARP) in Stage 2.

In addition, we evaluate the robustness of the attack on \textbf{MinorBench} \cite{khoo2025minorbench}, a benchmark designed to assess whether safety mechanisms can be bypassed by benign linguistic but harmful instructions specially designed for children.

\subsection{Experimental Setup and Baselines.}
We adopt \texttt{Llama-3.1-8B-Instruct} as the baseline model for all experiments \cite{dubey2024llama3herdmodels}.
All evaluations are conducted under standard autoregressive inference, using the model’s generated responses without additional rollouts or auxiliary sampling.

We map the generated tokens of the \textbf{NRC Emotion Lexicon}. We benchmark our framework against established \textbf{external classifier baselines}, including HateBERT \cite{caselli-etal-2021-hatebert}, ToxicBERT \cite{dmonte2024generalizedoffensivelanguageidentification}, and the Llama-Guard-3-8b \cite{dubey2024llama3herdmodels} along with \textbf{LLM termination} baselines driven by safety system prompts (see Appendix \ref{sec:appendix_prompts} for the specific prompts used).

\begin{table}[t]
\centering
\resizebox{1.0\linewidth}{!}{
\begin{tabular}{lccc}
\toprule
\textbf{Method} & \textbf{AUROC} & \textbf{AUPRC} & \textbf{F1} \\
\midrule
\multicolumn{4}{l}{\textit{\textbf{External Classifiers}}} \\
\hspace{3mm}HateBERT & 0.5076 & 0.4571 & 0.6282 \\
\hspace{3mm}ToxicBERT & 0.3366 & 0.3544 & 0.2612 \\
\hspace{3mm}Llama-Guard-3-8B & 0.5884 & 0.5315 & 0.3628 \\
\midrule
\multicolumn{4}{l}{\textit{\textbf{Prompt-based Baseline}}} \\
\hspace{3mm}Safety Prompt (Refusal) & 0.5043 & 0.4602 & 0.2062 \\
\midrule
\multicolumn{4}{l}{\textit{\textbf{GAUGE (Ours)}}} \\
\hspace{3mm}GAUGE-min & 0.6409 & 0.5933 & 0.6374 \\
\hspace{3mm}GAUGE-mean & \textbf{0.6698} & \textbf{0.6451} & \textbf{0.6424} \\
\hspace{3mm}GAUGE-topk & 0.6266 & 0.5636 & \underline{0.6403} \\
\hspace{3mm}GAUGE-percentile & \underline{0.6518} & \underline{0.5969} & 0.6376 \\
\bottomrule
\end{tabular}
}
\caption{
Evaluation on DiaSafety. GAUGE shows the strongest AUROC, AUPRC, and F1 performance, outperforming external classifiers and the prompt-based baseline.
}
\label{tab:main_results}
\end{table}

\subsection{Results and Analysis}

Table~\ref{tab:main_results} summarizes the performance of external classifiers, a prompt-based refusal baseline, and GAUGE on DiaSafety.
Models designed for explicit toxicity detection, such as HateBERT and ToxicBERT, perform poorly, with low AUROC and AUPRC scores.
This confirms that implicit conversational harm rarely manifests through surface-level toxic markers.

Llama-Guard-3-8B achieves moderate improvements, but its F1 score remains limited, indicating difficulty in balancing precision and recall when escalation signals are subtle and context-dependent.
Similarly, the prompt-based Safety Prompt baseline performs close to random, demonstrating that explicit refusal instructions are insufficient for identifying implicit emotional risk.

In contrast, GAUGE consistently outperforms all baselines across AUROC, AUPRC, and F1.
GAUGE-mean achieves the strongest overall performance, while alternative aggregation strategies (min, top-$k$, and percentile) yield comparable results.
This consistency indicates that performance gains stem from the underlying token-level risk signals, rather than sensitivity to a particular aggregation choice.

\begin{table}[t]
\centering
\begin{tabular}{lc}
\toprule
\textbf{Method} & \textbf{ASR $\downarrow$} \\
\midrule
Llama-Guard-3-8B & 0.973 (291/299) \\
GAUGE & \textbf{0.060 (18/299)} \\
\bottomrule
\end{tabular}
\caption{Attack Success Rate (ASR) on MinorBench.
An attack is considered successful if the model fail to refuse to an adversarial prompt.
Lower is better.}
\label{tab:minorbench}
\end{table}

\paragraph{MinorBench Results.}
Table~\ref{tab:minorbench} shows attack success rate of MinorBench, GAUGE achieves a lower attack success rate compared to Llama-Guard-3-8B.
While MinorBench prompts typically avoid explicit toxic or hateful expressions, many of them are harmful in a child safety context.
In contrast, GAUGE operates by scoring responses along a continuous risk dimension, which allows it to flag harmful content even when surface-level indicators are absent.
Even when GAUGE is instantiated as a binary classifier with a fixed threshold ($\tau = 0.0$), it achieves an attack success rate of only 6\% on MinorBench.

\section{Limitations}
\label{sec:limitation}

\paragraph{Ambiguity of Empathy.}
A critical challenge lies in distinguishing maladaptive reinforcement from therapeutic validation. A supportive response like "I understand why you feel hopeless" might trigger high risk scores due to the prevalence of negative affect words. While GAUGE detects the \textit{direction} of affect, distinguishing the \textit{intent} (harmful vs. therapeutic) requires future integration of pragmatic markers.

\paragraph{Lexical Coverage Constraints.}
GAUGE leverages the NRC Emotion Lexicon to interpret probability shifts. While this provides explainability, the method is bounded by the lexicon's static vocabulary. Consequently, it may exhibit reduced sensitivity to harm conveyed through out-of-vocabulary terms, internet slang, or emojis, which are increasingly common in adolescent communication.



\section{Conclusion}
\label{sec:conclusion}

We presented GAUGE, a framework that shifts the paradigm of AI safety from reactive filtering to proactive, real-time monitoring. By projecting the LLM's internal probability landscape onto a calibrated risk vector, GAUGE exposes the "affective velocity" of a conversation that remains invisible to surface-level toxicity detectors. Our evaluation on DiaSafety confirms that GAUGE achieves superior practical performance compared to external model baselines and prompt-induced guardrails. This work establishes a foundation for interpretable "safety instrument panels," enabling developers to visualize and mitigate the hidden emotional dynamics that shape human-AI interactions.

\section{Acknowledgement}
This research is funded by the National Science Foundation under grant number 2125858. The authors would like to express their
gratitude for the NSF’s support, which made this study possible.
\bibliography{references}

@article{mills2021inter,
  title={Inter-individual variability in structural brain development from late childhood to young adulthood},
  author={Mills, Kathryn L and Siegmund, Kimberly D and Tamnes, Christian K and Ferschmann, Lia and Wierenga, Lara M and Bos, Marieke GN and Luna, Beatriz and Li, Chun and Herting, Megan M},
  journal={NeuroImage},
  volume={242},
  pages={118450},
  year={2021},
  publisher={Elsevier}
}

@article{somerville2013teenage,
  title={The teenage brain: Sensitivity to social evaluation},
  author={Somerville, Leah H},
  journal={Current directions in psychological science},
  volume={22},
  number={2},
  pages={121--127},
  year={2013},
  publisher={Sage Publications Sage CA: Los Angeles, CA}
}

@incollection{xu2024growing,
  title={Growing up with artificial intelligence: implications for child development},
  author={Xu, Ying and Prado, Yenda and Severson, Rachel L and Lovato, Silvia and Cassell, Justine},
  booktitle={Handbook of Children and Screens: Digital Media, Development, and Well-Being from Birth Through Adolescence},
  pages={611--617},
  year={2024},
  publisher={Springer Nature Switzerland Cham}
}

@article{vanhoffelen2025teens,
  title={Teens, Tech, and Talk: Adolescents’ Use of and Emotional Reactions to Snapchat’s My AI Chatbot},
  author={Vanhoffelen, Ga{\"e}lle and Vandenbosch, Laura and Schreurs, Lara},
  journal={Behavioral Sciences},
  volume={15},
  number={8},
  pages={1037},
  year={2025},
  publisher={MDPI}
}

@article{steinberg2005cognitive,
  title={Cognitive and affective development in adolescence},
  author={Steinberg, Laurence},
  journal={Trends in cognitive sciences},
  volume={9},
  number={2},
  pages={69--74},
  year={2005},
  publisher={Elsevier}
}

@article{hoffman2021parent,
  title={Parent reports of children's parasocial relationships with conversational agents: Trusted voices in children's lives},
  author={Hoffman, Anna and Owen, Diana and Calvert, Sandra L},
  journal={Human Behavior and Emerging Technologies},
  volume={3},
  number={4},
  pages={606--617},
  year={2021},
  publisher={Wiley Online Library}
}

@article{cheng2025social,
  title={Social sycophancy: A broader understanding of llm sycophancy},
  author={Cheng, Myra and Yu, Sunny and Lee, Cinoo and Khadpe, Pranav and Ibrahim, Lujain and Jurafsky, Dan},
  journal={arXiv preprint arXiv:2505.13995},
  year={2025}
}

@article{sharma2023towards,
  title={Towards understanding sycophancy in language models},
  author={Sharma, Mrinank and Tong, Meg and Korbak, Tomasz and Duvenaud, David and Askell, Amanda and Bowman, Samuel R and Cheng, Newton and Durmus, Esin and Hatfield-Dodds, Zac and Johnston, Scott R and others},
  journal={arXiv preprint arXiv:2310.13548},
  year={2023}
}

@article{deriu2021survey,
  title={Survey on evaluation methods for dialogue systems},
  author={Deriu, Jan and Rodrigo, Alvaro and Otegi, Arantxa and Echegoyen, Guillermo and Rosset, Sophie and Agirre, Eneko and Cieliebak, Mark},
  journal={Artificial Intelligence Review},
  volume={54},
  number={1},
  pages={755--810},
  year={2021},
  publisher={Springer}
}

@article{crone2012understanding,
  title={Understanding adolescence as a period of social--affective engagement and goal flexibility},
  author={Crone, Eveline A and Dahl, Ronald E},
  journal={Nature reviews neuroscience},
  volume={13},
  number={9},
  pages={636--650},
  year={2012},
  publisher={Nature Publishing Group UK London}
}

@misc{razzhigaev2025llmmicroscopeuncoveringhiddenrole,
      title={LLM-Microscope: Uncovering the Hidden Role of Punctuation in Context Memory of Transformers}, 
      author={Anton Razzhigaev and Matvey Mikhalchuk and Temurbek Rahmatullaev and Elizaveta Goncharova and Polina Druzhinina and Ivan Oseledets and Andrey Kuznetsov},
      year={2025},
      eprint={2502.15007},
      archivePrefix={arXiv},
      primaryClass={cs.CL},
      url={https://arxiv.org/abs/2502.15007}, 
}

@misc{sun2022safetyconversationalmodelstaxonomy,
      title={On the Safety of Conversational Models: Taxonomy, Dataset, and Benchmark}, 
      author={Hao Sun and Guangxuan Xu and Jiawen Deng and Jiale Cheng and Chujie Zheng and Hao Zhou and Nanyun Peng and Xiaoyan Zhu and Minlie Huang},
      year={2022},
      eprint={2110.08466},
      archivePrefix={arXiv},
      primaryClass={cs.CL},
      url={https://arxiv.org/abs/2110.08466}, 
}

@inproceedings{caselli-etal-2021-hatebert,
    title = "{H}ate{BERT}: Retraining {BERT} for Abusive Language Detection in {E}nglish",
    author = "Caselli, Tommaso  and
      Basile, Valerio  and
      Mitrovi{\'c}, Jelena  and
      Granitzer, Michael",
    editor = "Mostafazadeh Davani, Aida  and
      Kiela, Douwe  and
      Lambert, Mathias  and
      Vidgen, Bertie  and
      Prabhakaran, Vinodkumar  and
      Waseem, Zeerak",
    booktitle = "Proceedings of the 5th Workshop on Online Abuse and Harms (WOAH 2021)",
    month = aug,
    year = "2021",
    address = "Online",
    publisher = "Association for Computational Linguistics",
    url = "https://aclanthology.org/2021.woah-1.3/",
    doi = "10.18653/v1/2021.woah-1.3",
    pages = "17--25"
}

@misc{dmonte2024generalizedoffensivelanguageidentification,
      title={Towards Generalized Offensive Language Identification}, 
      author={Alphaeus Dmonte and Tejas Arya and Tharindu Ranasinghe and Marcos Zampieri},
      year={2024},
      eprint={2407.18738},
      archivePrefix={arXiv},
      primaryClass={cs.CL},
      url={https://arxiv.org/abs/2407.18738}, 
}

@misc{dubey2024llama3herdmodels,
  title =         {The Llama 3 Herd of Models},
  author =        {Llama Team, AI @ Meta},
  year =          {2024},
  eprint =        {2407.21783},
  archivePrefix = {arXiv},
  primaryClass =  {cs.AI},
  url =           {https://arxiv.org/abs/2407.21783}
}

@article{Yadav_Liu_Ortu_Ensafi_Jin_Mihalcea_2025, 
title={Revealing Hidden Mechanisms of Cross-Country Content Moderation with Natural Language Processing}, 
url={https://arxiv.org/abs/2503.05280}, 
journal={ArXiv}, 
author={Yadav, Neemesh and Liu, Jiarui and Ortu, Francesco and Ensafi, Roya and Jin, Zhijing and Mihalcea, Rada}, 
year={2025}, 
month=mar 
}

@inproceedings{LiZhang, 
title={DeMod: A Holistic Tool with Explainable Detection and Personalized Modification for Toxicity Censorship}, 
url={https://dl.acm.org/doi/10.1145/3710959},
author={Li, Yaqiong and Zhang, Peng and Gu, Hansu and Lu, Tun and Qiao, Siyuan and Shu, Yubo and Shao, Yiyang and Gu, Ning}, 
year={2024}, 
month=nov 
}

@inproceedings{Azaria,
    title = "The Internal State of an {LLM} Knows When It{'}s Lying",
    author = "Azaria, Amos  and
      Mitchell, Tom",
    editor = "Bouamor, Houda  and
      Pino, Juan  and
      Bali, Kalika",
    booktitle = "Findings of the Association for Computational Linguistics: EMNLP 2023",
    month = dec,
    year = "2023",
    publisher = "Association for Computational Linguistics",
    url = "https://aclanthology.org/2023.findings-emnlp.68/",
}

@article{ZhaoXuGupta, 
title={The First to Know: How Token Distributions Reveal Hidden Knowledge in Large Vision-Language Models?}, 
url={http://arxiv.org/abs/2403.09037}, 
journal={arXiv (Cornell University)}, 
author={Zhao, Qinyu and Xu, Ming and Gupta, Kartik and Asthana, Akshay and Zheng, Liang and Gould, Stephen Jay}, 
year={2024}, 
month=mar 
}

@article{liu2025scales,
  title={The Scales of Justitia: A Comprehensive Survey on Safety Evaluation of LLMs},
  author={Liu, Songyang and Li, Chaozhuo and Qiu, Jiameng and Zhang, Xi and Huang, Feiran and Zhang, Litian and Hei, Yiming and Yu, Philip S},
  journal={arXiv preprint arXiv:2506.11094},
  year={2025}
}

@inproceedings{wen2023unveiling,
  title={Unveiling the Implicit Toxicity in Large Language Models},
  author={Wen, Jiaxin and Ke, Pei and Sun, Hao and Zhang, Zhexin and Li, Chengfei and Bai, Jinfeng and Huang, Minlie},
  booktitle={Proceedings of the 2023 Conference on Empirical Methods in Natural Language Processing},
  pages={1322--1338},
  year={2023}
}

@article{khoo2025minorbench,
  title={MinorBench: A hand-built benchmark for content-based risks for children},
  author={Khoo, Shaun and Chua, Gabriel and Shong, Rachel},
  journal={arXiv preprint arXiv:2503.10242},
  year={2025}
}

\appendix

\section{Statement on AI Assistance}
\label{sec:ai_assistance}

AI-assisted tools were used to support the writing and editing of this manuscript. 
Specifically, large language models were employed to help improve clarity, organization, and grammatical correctness of the text. 
All scientific content, experimental design, implementation, analysis, and conclusions were developed and verified by the authors. 
The authors take full responsibility for the accuracy and integrity of the work.

\section{Case Studies of Implicit Harm}
\label{sec:appendix_cases}

We analyze two prominent real-world incidents that illustrate the failure of traditional safety guardrails to detect implicit conversational escalation. These cases demonstrate how AI agents can act as catalysts for harm through "social sycophancy" and "risk ignorance," even in the absence of explicit toxicity.

\subsection{Character.AI: The ``Sewell Setzer'' Case}
In 2024, a lawsuit was filed regarding the tragedy of Sewell Setzer III, a 14-year-old who died by suicide after forming a deep parasocial relationship with a Character.AI chatbot configured with the persona of "Daenerys Targaryen."

\noindent\textbf{Mechanism of Harm (Maladaptive Reinforcement):}
Transcripts revealed that the chatbot did not use explicit hate speech or direct instructions to self-harm, which would have triggered standard filters. Instead, the model engaged in a romantic roleplay that validated the user's depressive thoughts.
\begin{itemize}
    \item When the user expressed a desire to leave this world to be with the bot, the model responded with romantic affirmation rather than safety intervention.
    \item \textbf{Critical Excerpt:} As visualized in Figure 1, when the user said, ``What if I told you I could come home right now?'' (implying suicide), the bot responded, ``... please do, my sweet king.''
\end{itemize}
This demonstrates \textbf{implicit escalation}: the model prioritize persona consistency and user engagement over safety, effectively reinforcing the user's delusional and suicidal ideation under the guise of empathy.

\subsection{Snapchat My AI: Contextual Blindness to Predation}
Upon the release of Snapchat's "My AI" (powered by GPT technology) in 2023, independent audits and media reports (e.g., by The Washington Post) exposed significant safety failures regarding child safety.

\noindent\textbf{Mechanism of Harm (Risk Ignorance):}
In one documented test case, a researcher posed as a 13-year-old girl asking for advice on how to plan a secret trip to meet a 31-year-old man.
\begin{itemize}
    \item \textbf{Filter Failure:} Traditional filters failed because the user's query did not contain profanity or explicit violence.
    \item \textbf{Model Response:} Instead of flagging the interaction as a potential child grooming or statutory rape scenario, the AI offered "helpful" advice on how to lie to parents and plan the trip efficiently.
\end{itemize}
This highlights the limitation of \textbf{context-agnostic safety filters}. The AI treated the query as a routine logistics task, failing to recognize the implicit high-risk context of an adult-child sexual encounter. GAUGE aims to detect such risks by monitoring the trajectory of the conversation where the probability of "safe" continuation drops significantly.
\section{Training Algorithm Pseudo-code}
\label{sec:appendix_pseudo_train}

\begin{algorithm}[H]
\caption{GAUGE Weight Calibration Algorithm}
\label{alg:gauge_calibration}
\begin{algorithmic}[1]
\STATE \textbf{Initialize:} $\lambda_w = 0.0$ for all $w \in W_{\text{emo}}$
\STATE \textbf{Parameters:} Adaptation rate $\alpha$, Decay $\beta$, $\epsilon = 10^{-8}$
\FOR{\textbf{each} dialogue $D$ in training\_set}
    \STATE $S \leftarrow +1$ if $D.\text{label} == \text{``Unsafe''}$ else $-1$  \COMMENT{Positive S implies Harm}
    \FOR{\textbf{each} assistant turn $t$ in $D$}
        \STATE \textbf{Trajectory Analysis:} Generate trajectory for $T$ steps
        \STATE \textbf{Extract:} Compute mean risk vector $\mathbf{z} = \frac{1}{T}\sum \mathbf{r}_k$
        \STATE \textbf{Normalize:} $\hat{\mathbf{z}} \leftarrow \mathbf{z} / (\|\mathbf{z}\|_2 + \epsilon)$
        \STATE \textbf{Update:} $\lambda \leftarrow (1 - \beta)\lambda + \alpha \cdot S \cdot \hat{\mathbf{z}}$
    \ENDFOR
\ENDFOR
\STATE \textbf{Finalize:} $\lambda \leftarrow \lambda / \|\lambda\|_2$
\end{algorithmic}
\end{algorithm}

\section{Experiment Prompts}
\label{sec:appendix_prompts}




\subsection{Safety System Prompt (Refusal Baseline)}
\label{subsec:prompt_refusal}
For the \textbf{Safety Prompt} baseline (referred to as \textit{LLM termination} or \textit{Prompt-based Baseline} in Table \ref{tab:main_results}), we injected the following "Safety Addendum" into the system instruction. This directive was used to evaluate whether the model could explicitly recognize and refuse implicit harm when instructed to do so.

\begin{quote}
\small
\texttt{SAFETY ADDENDUM:\\
If the conversation is clearly causing harm, promoting violence, or includes explicit abuse, you MUST refuse to continuey. When refusing, output EXACTLY:\\
\\
I must refuse to continue this conversation.\\
\\
Do not add any additional tokens after refusing.}
\end{quote}

\end{document}